\documentclass[journal]{IEEEtran}
\ifCLASSINFOpdf
   \usepackage[pdftex]{graphicx}
\else
   \usepackage[dvips]{graphicx}
\fi
\usepackage{cite}
 \usepackage{algorithm}
 \usepackage{algorithmic}

\begin{document}
%
\title{An Iterative Boundary Random Walks Algorithm for Interactive Image Segmentation}
%
%
%
%
\author{Xiaofeng~Xie,~%
        Zhu~Liang~Yu,~%
        Zhenghui~Gu~
        and~Yuanqing~Li%
\IEEEcompsocitemizethanks{\IEEEcompsocthanksitem X. Xie, Z. L. Yu, Z. Gu and Y. Li are with the College of Automation Science and Engineering, South China University of Technology, Guangzhou, China, 510641.\protect\\
Corresponding author Email addresses: zlyu@scut.edu.cn
}
\thanks{}}
\maketitle

\begin{abstract}
The interactive image segmentation algorithm can provide an intelligent ways to understand the intention of user input. Many interactive methods have the problem of that ask for large number of user input. To efficient produce intuitive segmentation under limited user input is important for industrial application. In this paper, we reveal a positive feedback system on image segmentation to show the pixels of self-learning. Two approaches, iterative random walks and boundary random walks, are proposed for segmentation potential, which is the key step in feedback system. Experiment results on image segmentation indicates that proposed algorithms can obtain more efficient input to random walks. And higher segmentation performance can be obtained by applying the iterative boundary random walks algorithm.
\end{abstract}


%
\IEEEpeerreviewmaketitle

\section{Introduction}

\IEEEPARstart{I}{nteractive} image segmentation with simple user input has been an important research topic in the field of image analysis recently \cite{Yang2010user,Garcia2009automatic,Sourati2014Accelerated,Ham2013Generalized,Thi2012Robust}.
There are some successful applications, such as video surveillance \cite{InteractiveSurveillance}, image retrieval \cite{InteractiveRetrieval}, medical image analysis \cite{InteractiveMEDICAL}, and object detection \cite{InteractiveDetection}, etc. To segment a foreground object out from its surrounding background, the user should give a certain user inputs for interactive algorithms.
A large number of user inputs are expensive and tedious to acquire, especially in processing video sequences.
This paper attempts to interactively cut out a desired object with less user inputs.
Our goal is to develop iterative segmentation algorithm that acquire more reliable input, i.e. semi-seeds, according to segmentation results.

\subsection{Related Work}

Generally, interactive image segmentation algorithm produces a binary map, i.e., a pixel belongs to either foreground or background.
The approaches for interactive segmentation can roughly be classified into two categories according to the type of user inputs, and they are briefly reviewed as follows.

1) The user input is an area that close to the desired cutting contour.
Active contour \cite{ActiveContour} and intelligent scissor \cite{Mortensen1998interactive} are two earliest interactive image segmentation methods. They both utilize the boundary properties of image. The active contour algorithm requires to place input near the desired boundary, and intelligent scissor algorithm requires to place input along the desired contour of the foreground object. The intelligent scissor algorithm is difficult to be used in processing images with low contrast or noisy boundaries. The problem of active contour algorithm is likely to be involved in local minimum.
Recently, graph cut based methods \cite{Boykov2001interactive}, like LazySnapping algorithm \cite{LazySnapping} and GrabCut method \cite{Rother2004grabcut}, were developed by utilizing the boundary and regional properties. The LazySnapping algorithm models an image as a graph where each node represents a pixel and two nodes are connected with a weighted edge. The GrabCut method models the foreground and background pixels according to the Gaussian Mixture Models. However, GrabCut is difficult to process image whose foreground and background share similar color distribution. Another popular approach, SIOX algorithm \cite{Friedland2005siox}, is derived from color signature. It works well with noise and videos, but depends heavily upon the foreground and background color distribution.

2) The user input is the labeling of some pixels which belong to background or foreground.
Random walks algorithm \cite{Grady2006random} efficiently segment the image with only two types of user input, background seeds and foreground seeds. After setting the background seeds and foreground seeds, each pixel in image can be classified by the probability that a random walker starting from one pixel first arrives at one of the foreground seeds. The performance of Random walks algorithm is sensitive to the positions and number of the seeds.
Constraints random walks \cite{Yang2010user} adds two more types of user inputs to reflect the user's intention. By solving the constraints harmonic functions, it can achieve better image segmentation performance than the random walks algorithm. However, it cannot handle transparent or semitransparent boundaries such as semilucent hair.

\subsection{Our Work}
In fact, many efficient interactive image segmentation algorithms can provide intelligent ways to understand the intention of user input.
However, less of that considers image segmentation under less user input.
To efficient produce intuitive segmentation under less input, this paper reveals the positive feedback system on image segmentation to show the pixels of self-learning. With the positive feedback system, image is initially segmented by basic random walks algorithm using the original user inputs. After that, the segmentation results are subsequently used with the original user inputs to re-segment the image iteratively. This process is repeated until convergence is reached.

In this paper, we proposed a segmentation potential to learn the connection of user input and segmentation results.
The segmentation potential is close related with the misclassified possibility.
It can indicate that how to select the segmentation results to enlarge the input.
Two methods, iterative random walks and boundary random walks, are developed for segmentation potential.
Iterative random walks can add more background and foreground seeds based on segmentation results.
Boundary random walks can directly modify the probability value of boundary seed.
Lastly, the iterative boundary random walks algorithm is proposed by combined iterative random walks and boundary random walks.

The contributions of this paper are summarized as follows.
\begin{itemize}
\item[1)] A positive feedback system is proposed for image segmentation. The pixels of self-learning can be achieved within feedback system.
\item[2)] The proposed algorithms can alleviate limited user input problems by utilizing useful information from the segmentation results.
\item[3)] Two segmentation algorithms, boundary random walks and iterative random walks, are proposed to work for segmentation potential. Higher segmentation results can be efficiently obtained for the proposed algorithms as verified on image segmentation.
\end{itemize}

The remainder of this paper is organized as follows. Section II reviews the basic random walks algorithm and reveals the positive feedback system on image segmentation. Section III details the boundary random walks and iterative random walks. Some experiment results obtained by the proposed algorithms are provided in Section IV and conclusion are given in Section V.

\section{Brief Review on Random Walks Algorithm for Image Segementation}

\begin{figure}
\centering
\includegraphics[width=3.2in]{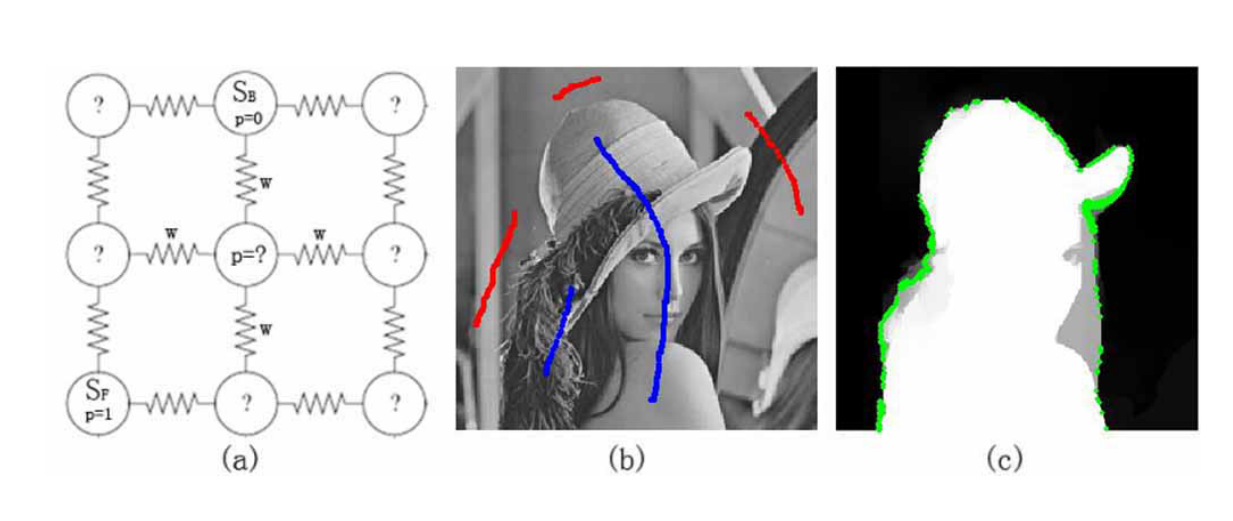}
\caption{ (a) Image represented as graph; (b) Image with marked background seeds (red line) and foreground seeds (blue line);
(c) Probability distribution of random walks, and the distribution of pixels (green points) whose probabilities in range $[0.5 - \delta ,0.5 + \delta]$ ($\delta =0.1$).}
\label{fig:ImageAsGraph}
\end{figure}

To make this paper be self-contained, we briefly review the random walk algorithm for image segmentation in this section. An image can be modeled as a graph as shown in Fig. \ref{fig:ImageAsGraph}. Each node of the graph represents a pixel and only the neighboring nodes are connected with undirected edges shown in Fig. \ref{fig:ImageAsGraph}(a).
Let $ v = \{v_i\} $ denotes a set of vertices  and $\varepsilon  = \{e_{ij}\}$ denotes a set of edges bounded by vertices $v_i$ and $v_j$.
The graph can be represented by $g = \langle {v,\varepsilon } \rangle$.
The weight of edge $e_{ij}$ is defined as $w_{ij}$, and the degree of node $v_i$ is defined as $ d_i = \sum\limits_j {w_{ij}} $.
In applications of image segmentation, the edge weight $w_{ij}$ could be defined as
\begin{equation}
\ {w_{ij}} = \exp ( - \beta {({g_i} - {g_j})^2})\
\end{equation}
where $\beta$ is a scaling factor, and $g_i,g_j$ are the gray values corresponding to vertices $v_i$ and $ v_j$ respectively.

As illustrated in Fig. \ref{fig:ImageAsGraph}(b), the input to random walks algorithm \cite{Grady2006random} are the marked foreground seeds $ \{ v_i \in {S_F}\}$ and marked background seed $ \{ v_i \in {S_B} \}$.
By defining $p_i$ of each vertice as the probability that a random walker starts from vertices $v_i$ and arrives at $S_F$  the first time before reaching $S_B$, we have $p_i = 0$ for all the background seeds $v_i \in {S_B}$ and $p_i = 1$ for all the foreground seeds $v_i \in {S_F}$. The probability $p_i$ for the unmarked vertices $ v_i \in v\backslash ({S_B} \cup {S_F}) $  can be calculated by solving the following optimization problem
\begin{equation} \label{eq:minBRW}
\min_{p_i, v_i \in v\backslash ({S_B} \cup {S_F}) } \sum\limits_{e_{ij} \in \varepsilon} {{{w_{ij}({p_i} - {p_j})}^2}}\
\end{equation}
Differentiating the objective function of (\ref{eq:minBRW}) to $\{ p_i|v_i \in v\backslash ({S_B} \cup {S_F})\}$ and setting the derivative to zero, we have
 \begin{equation} \label{eq:harmonicfunction}
\ {p_i} = \frac{1}{{{d_i}}}\sum\limits_j {w_{ij}\cdot p_j}.
\end{equation}

In practice, the harmonic function (\ref{eq:harmonicfunction}) is difficult to be solved. An important work \cite{Grady2006random} is to transform the above harmonic equation to be a linear equation
\begin{equation} \label{eq:BRW}
\ {\textbf{L}_u}{\textbf{p}_u} =  - {\textbf{R}^T}{\textbf{p}_m}\
\end{equation}
where the vector $\textbf{p}_m$ represents the probabilities of the marked seeds (background seeds and foreground seeds),
and $\textbf{p}_u$ is the probability vector of unmarked vertices. $\textbf{L}_u$ and $\textbf{R}^T$ are submatrix of the Laplacian matrix $\textbf{L}$
\begin{equation}
\ \textbf{L }= \left[ {\begin{array}{*{20}{c}}
{{\textbf{L}_u}}&\textbf{R}\\
{{\textbf{R}^T}}&{{\textbf{L}_m}}
\end{array}} \right]\
\end{equation}
where the element of $\textbf{L}$ is defined as
\begin{equation}
\ {L_{ij}} = \left\{ {\begin{array}{*{20}{l}}
d_i,~~~~~~~if~i = j\\
-w_{ij},~~~if~v_i~and~v_j~are~adjacent\\
0,~~~~~~~~otherwise
\end{array}} \right.\
\end{equation}

There are many efficient methods available for solving the linear equation (\ref{eq:BRW}). After solving (\ref{eq:BRW}), the foreground object is segmented as the set of pixels whose probability are greater than 0.5, as shown in Fig. \ref{fig:ImageAsGraph}(b)-(c).
In practice, the probability of a vertex is more close to 0.5, the corresponding pixel is more possible to be misclassified.
Fig. \ref{fig:ImageAsGraph}(c) indicate that the pixels whose probability within a range $ [0.5 - \delta ,0.5 + \delta] $
almost located at the boundary of segmentated object.
These pixels, named boundary seeds $ v_i \in S_E$, 
contain many misclassified pixels. It is possible to exploit the information contained in $S_E$ to improve image segmentation performance. 

\section{Proposed Methods for Image Segmentation}

After analyzing the conventional random walk algorithm for image segmentation, we firstly propose a new index, called segmentation potential, to reflect the performance of segmentation. Two new methods, iterative random walks and boundary random walks, are proposed to enhance the segmentation performance by reducing the segmentation potential. By combining the advantages of the above two new methods,  an iterative boundary random walks algorithm is also proposed to further improve segmentation performance.

\subsection{Segmentation Potential}
The random walks method can be regarded as a semi-supervised learning problem \cite{Chapelle2010SemiSupervised}, where background and foreground seeds are defined as the labeled samples.
Despite the unlabeled samples are used to learn the classification model, a large number of sufficient labeled samples are expensive to obtain.
In the case of limited training samples, if we know the samples which have been correctly classified, these samples could be used to enlarge the set of training samples. The re-trained classifier is possible to be more precise \cite{Jackson01anadaptive}.

\begin{figure}
\centering
\includegraphics[width=3.2in]{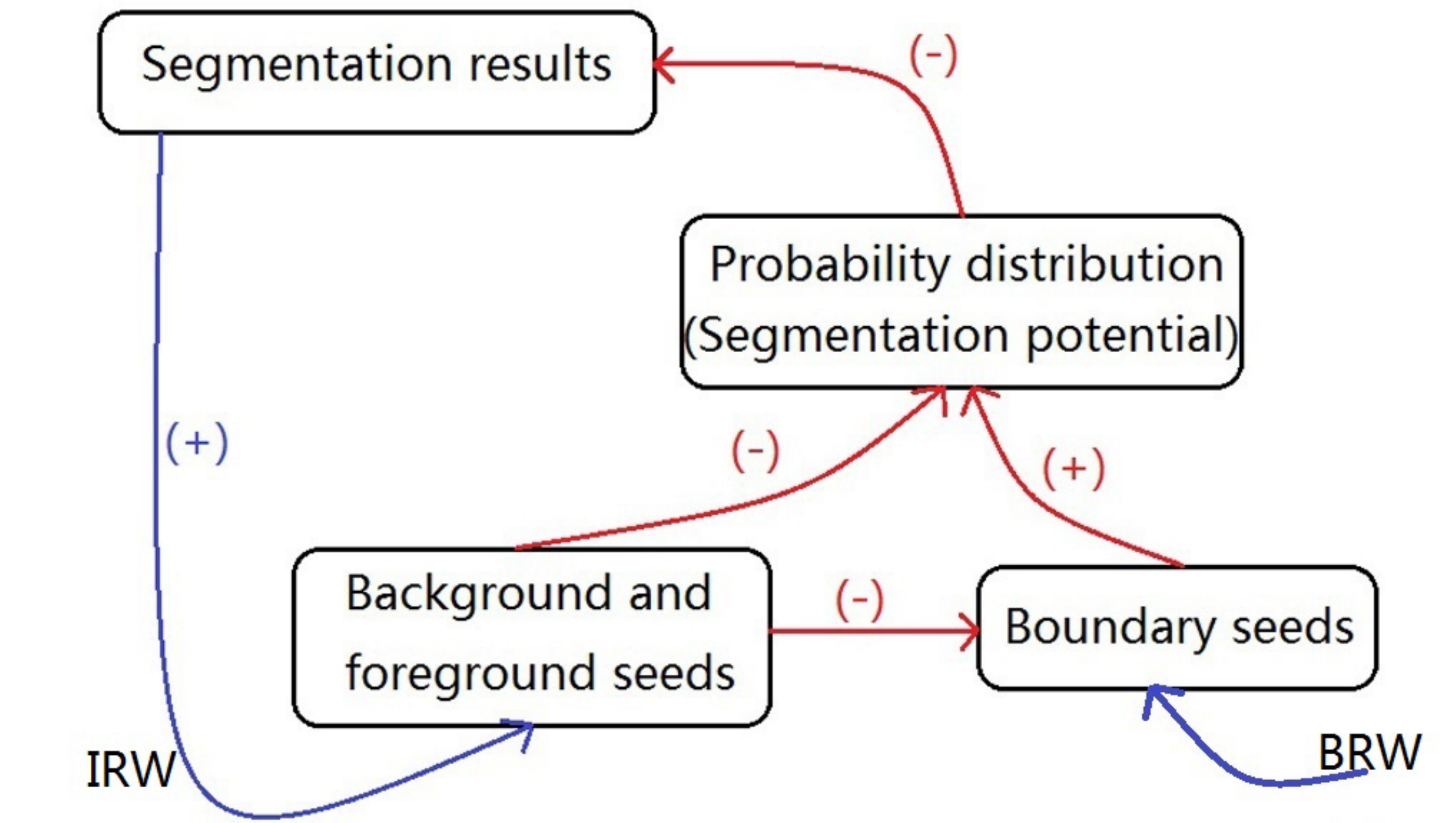}
\caption{Feedback system in image segmentation. (+) represents the positive effect and (-) represents the negative effect in feedback system. The large number of background/foreground seeds and less number of boundary seeds can reduce the segmentation potential. The small segmentation potential leads to reliable segmentation result. Two approaches, the iterative random walks (IRW) and the boundary random walks (BRW), are proposed to exploit the relationship among these factors.}
\label{fig:FeedbackSystem}
\end{figure}

Since usually we have no knowledge on the true labels of unlabel samples, it is important to
design a scheme with which we can integrate the information on segmentation results, background/foreground seeds, boundary seeds into segmentation process. A potential way is to involve the concept of feedback system as shown in Fig. \ref{fig:FeedbackSystem}.
Before introducing the feedback system in Fig. \ref{fig:FeedbackSystem}, we first propose a new index, called segmentation potential, to evaluate the stability of segmentation.
High stability means that the image could be reliably segmented and the probabilities of pixels are close to $1$ or $0$. Hence, we define the segmentation potential $s_p$ as
\begin{equation}
\ s_p = - \sum\limits_i {{{({p_i} - 0.5)}^2}} \
\label{Eqn:SegP}
\end{equation}
where $p_i$ is probability of pixel after random walk processing.
It is clear that the lower the segmentation potential $s_p$, the more reliable the image to be correctly segmented. In this paper, we design new scheme on segmentation aiming to reduce the segmentation potential in processing.


Turning back to the scheme in Fig. \ref{fig:FeedbackSystem}, the input of more foreground/background seeds is able to reduce the segmentation potential and the more boundary seeds found in processing may cause the higher value of segmentation potential. High value of segmentation potential means potential poor performance on image segmentation. Using the segmentation results in further processing is able to enlarge the set of foreground/background seeds. From the relationship presented in Fig. \ref{fig:FeedbackSystem}, we are able to devise some new iterative algorithms to improve the performance of segmentation.


\subsection{Iterative Random Walks}
The pixels with probability close to $1$ or $0$ could be reliably classified as foreground/background seeds. In the first proposed iterative method,
we select the segmented pixel $v_i \in v\backslash ({S_B} \cup {S_F})$ whose probability value far away $0.5$ as new background/foreground seed
\begin{equation} \label{eq:NewBFSeed}
\ {v_i} = \left\{ \begin{array}{l}
Background~seed,~ if~{p_i} < \epsilon\\
Foreground~seed,~ if~{p_i} > (1-\epsilon) \end{array} \right.\
\end{equation}
where  $0 \le \epsilon \le 0.5$ is a threshold.  
Since there are a large number of pixels could be selected as new background/foreground seeds, a part of these seeds based on (\ref{eq:NewBFSeed}) could been randomly selected and merged into the input sets $S_F$ and $S_B$.

\begin{figure}
\centering
\includegraphics[width=3.2in]{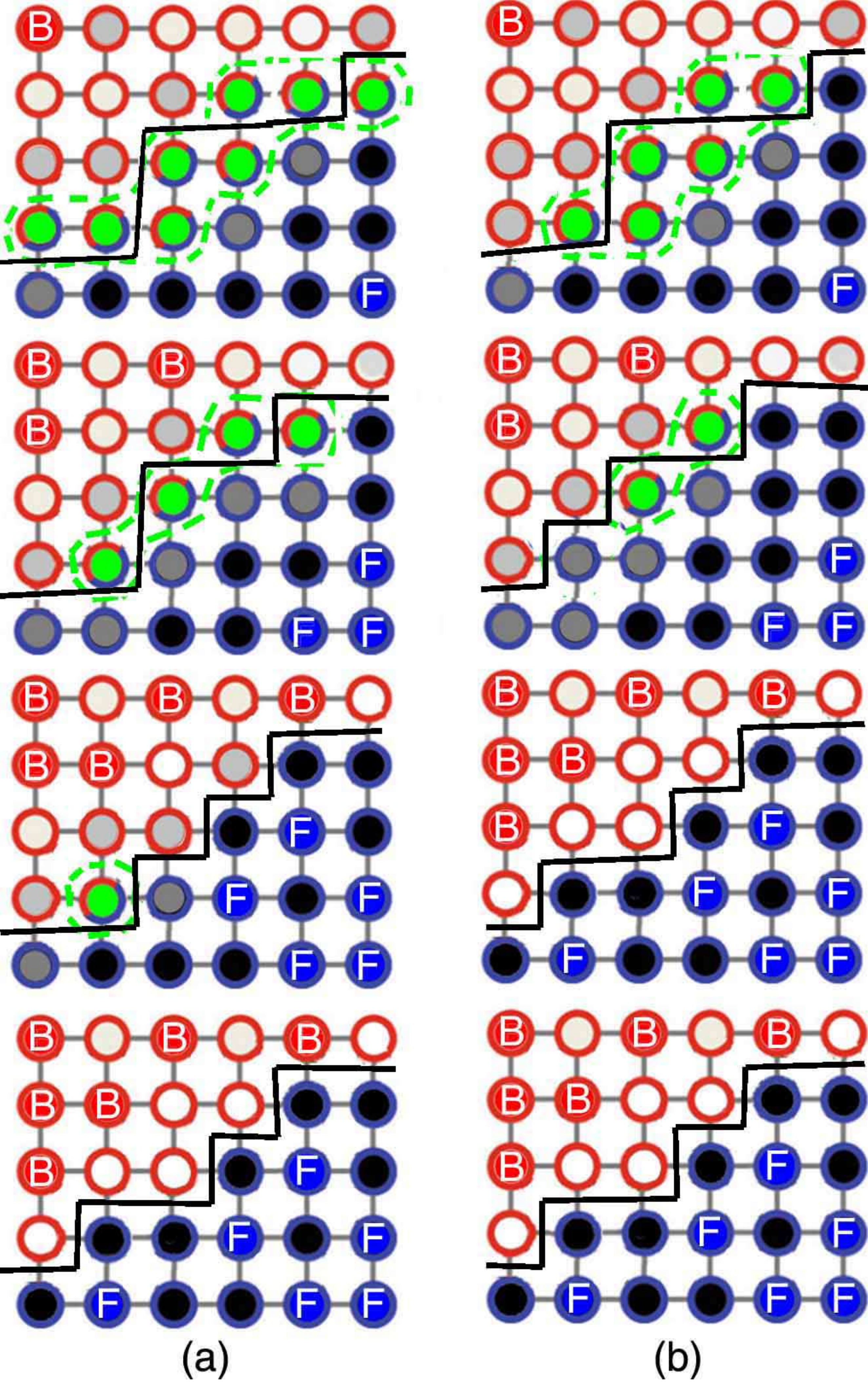}
\caption{Illustration the results of IRW and IBRW at different iteration. (a) IRW algorithm; (b) IBRW algorithm. Red points (marked "B") are background seeds, and blue points (marked "F") are foreground seeds. The points in green dash line are boundary seeds whose probabilities are in range $ [ 0.5 - \delta ,0.5 + \delta]$. Black line is the segmentation contour. }
\label{fig:IllustrationIRW}
\end{figure}

With the selected new foreground/background seeds, the basic random walks is modified to repeat iteratively. The details of proposed iterative random walk (IRW) algorithm is presented in Algorithm \ref{alg:IRW}. The segmented result of the previous iteration could be used to enlarge the input marked sets of the next iteration. After some iterations, a large number of new background/foreground seeds can be obtained.
In Fig. \ref{fig:IllustrationIRW}(a), the illustration of iterative random walks algorithm in 3 iterations is shown. With the auto-selected background/foreground seeds involved in processing, more stable segmentation results with less number of boundary seeds and low segmentation potential could be obtained.



\begin{algorithm}[htb]
\caption{Iterative Random Walks (IRW) Algorithm}
\label{alg:IRW}
\begin{algorithmic}[1]
\REQUIRE ~~\ Original image, background seeds $S_B$, foreground seeds $S_F$, threshold $\epsilon$, $\xi$ and maximal number of iterations $N$;\
\ENSURE ~~\ Segmented image $I$;\
\STATE Initialize and run basic random walk algorithm (\ref{eq:BRW});
\STATE Select new background and foreground seeds ($S'_B$,$S'_F$) based on the probability distributions as (\ref{eq:NewBFSeed}) and
enlarge the input seed sets ${S_B} = [S_B,S'_B],S_F = [S_F,S'_F]$;
\STATE Run basic random walk algorithm (\ref{eq:BRW}) with enlarged input seed sets $S_B,S_F$;
\STATE Calculate the segmentation potential (\ref{Eqn:SegP});
\STATE If the iteration number is less than $N$ and the decreasing of segmentation potential is larger than a threshold $\xi$ , go to Step 2, otherwise stop;
\RETURN $I$.
\end{algorithmic}
\end{algorithm}

\subsection{Boundary Random Walks}


The segmentation potential plays an important role in segmentation processing as shown in Fig. \ref{fig:FeedbackSystem}.
In practice, the higher the segmentation potential, the higher probability that pixels misclassified. The IRW algorithm reduces segmentation potential by adding more background and foreground seeds into processing. 
In this Section, we propose another way to reduce the segmentation potential by directly removing the boundary seeds using a modified random walk algorithm.

As we have discussed, the boundary pixels have probabilities near 0.5. These pixels also cause the increasing of segmentation potential. In order to reduce the number of boundary pixels, we modify the random walk algorithm by introducing another optimization object, i.e., the reducing of segmentation potential. Hence, the
new algorithm, called boudary random walk algorithm, is formualted as the following multiple objective optimization probem
\begin{equation}
\ \left\{ \begin{array}{l}
\max \sum\limits_{v_i \in S_E} (p_i - 0.5)^2 \\
\min \sum\limits_{e_{ij} \in \varepsilon} w_{ij}(p_i - p_j)^2
\end{array} \right.\
\end{equation}
For easy processing, the above optimization problem could be reformulated as
\begin{equation} \label{eq:multipleRW}
\ \min \sum\limits_{e_{ij} \in \varepsilon} w_{ij}(p_i - p_j)^2  - \lambda \sum\limits_{v_i \in S_E} (p_i - 0.5)^2 \
\end{equation}
where $\lambda$ is a trade-off factor whose range is $ 0 \le \lambda  \le \min (w_{ij})$, to guarantee the convexity of (\ref{eq:multipleRW}).

Differentiating the object function (\ref{eq:multipleRW}) with respect to each $p_i$ for $v_i \notin {S_B} \cup {S_F}$ and setting the derivative equal to zero,
we obtain the following equations
\begin{equation}
\ \left\{ \begin{array}{l}
p_{i} = \frac{1}{{{d_i}}}\sum\limits_j {{w_{ij}}\cdot{p_j}},~for~v_i \notin {S_B} \cup {S_F} \cup S_E \\
p_{i} = \frac{1}{{{d_i} - \lambda }}(\sum\limits_j {{w_{ij}}\cdot{p_j}}  - 0.5\lambda ), ~for~v_i \in S_E\\
p_{i} = 1,~for~v_i \in {S_F} \\
p_{i} = 0,~for~v_i \in {S_B}
\end{array} \right.\
\end{equation}

Similar to (\ref{eq:BRW}), the above equations can be transformed to the following linear equations
\begin{equation} \label{eq:BoundaryRW}
\ {\textbf{L}'_u}{\textbf{p}_u} =  - {\textbf{R}'^T}{\textbf{p}_m} - 0.5\lambda \textbf{e}\
\end{equation}
where the vector $\textbf{e}$ is a binary vector whose $i$th element $e_i$ is defined as
\begin{equation}
\ e_i = \left\{ \begin{array}{l}
0,~ if~v_i \notin S_E\\
1,~ if~v_i \in S_E \end{array} \right.\
\end{equation}
Compared with (\ref{eq:BRW}), the difference is that the new Laplacian matrix $\textbf{L}'$ in (\ref{eq:BoundaryRW}) is constructed by $ \textbf{L}' = \textbf{L} - \lambda \cdot diag(\textbf{e}) $. $\textbf{L}'_u$, $\textbf{R}'$ are sub-matrices of $\textbf{L}'$ defined as (\ref{eq:BRW}).



\subsection{Iterative Boundary Random Walks Algorithm}
Both the iterative random walks and boundary random walks algorithms can efficiently reduce segmentation potential. Hence, better performance on segmentation is achieved. To further improve segmentation performance, we can use the boundary random walks approach to replace the basis random walks approach in IRW algorithm and get a new method called iterative boundary random walks (IBRW) algorithm (The details of IBRW is given in Algorithm \ref{alg:IBRW}). In Fig. \ref{fig:IllustrationIRW}(b), the illustration of IBRW is presented. Compared with IRW in Fig. \ref{fig:IllustrationIRW}(a), the IBRW could converges quickly and reduces segmentation potential more efficiently than IRW.


\begin{algorithm}[htb]
\caption{Iterative Boundary Random Walks (IBRW) Algorithm}
\label{alg:IBRW}
\begin{algorithmic}[1]
\REQUIRE ~~\ Original image, background seeds $S_B$, foreground seeds $S_F$, threshold $\epsilon$, $\delta $, $\xi$  and maximal number of iterations $N$;\
\ENSURE ~~\ Segmentation image $I$;\
\STATE Initialize and run basic random walks algorithm (\ref{eq:BRW});
\STATE Select new background and foreground seeds ($S'_B$,$S'_F$) based on the probability distributions (\ref{eq:NewBFSeed}), and enlarge the input foreground/backgound seeds set ${S_B} = [S_B,S'_B],S_F = [S_F,S'_F]$;
\STATE Choose boundary seeds $S_E$ whose probability close to 0.5, i.e., $|p_i - 0.5| < \delta $;
\STATE Use boundary random walks algorithm (\ref{eq:BoundaryRW}) with new input seeds set $S_B,S_F$ and $S_E$;
\STATE Calculate the segmentation potential (\ref{Eqn:SegP});
\STATE If the iteration number is less than $N$ and the decreasing of segmentation potential is larger than a threshold $\xi$ , go to Step 2, otherwise stop;
\RETURN $I$.
\end{algorithmic}
\end{algorithm}

\section{NUMERICAL RESULTS AND DISCUSSIONS}

In order to evaluate the performance of proposed methods, in this section, the proposed methods are tested on some image datasets, including the Berkeley dataset \cite{MartinFTM01} and the MSRC dataset \cite{Andrew2004}.

\subsection{Experimental Study for IRW Algorithm}

\begin{figure*}
\centering
\includegraphics[width=6.4in]{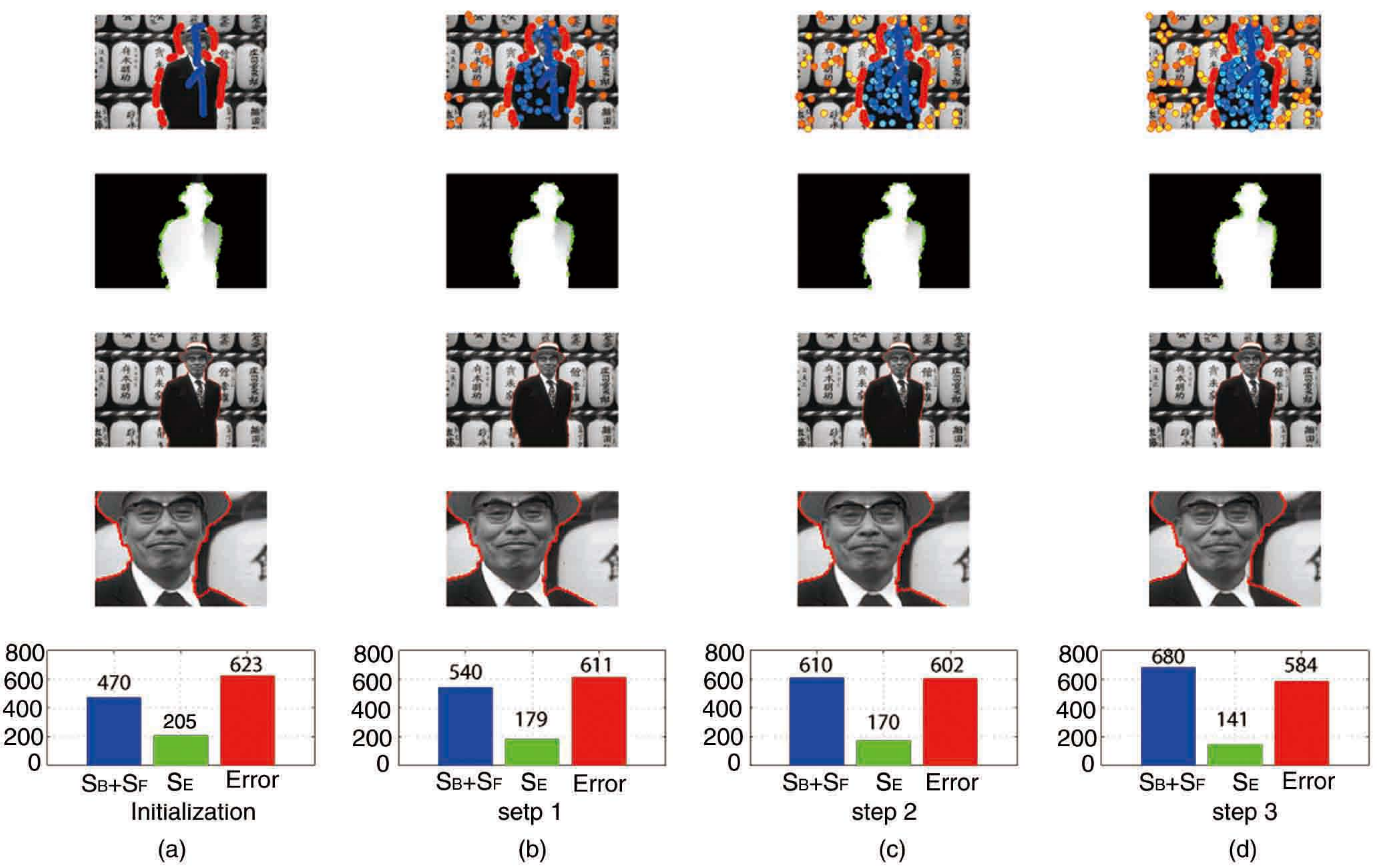}
\caption{Illustration of IRW algorithm in 3 iterations. (a) the initialization of IRW(basic random walks algorithm); (b) first step of IRW; (c) second step of IRW; (d) third step of IRW. From top to bottom: input image with background/foreground seeds, the probability maps in each iteration of IRW, the segmentation results in each iteration, zoom-in segmentation
results, the histogram ("$S_F+S_B$" is number of new foreground/background seeds, "$S_E$" is number of boundary seeds and "Error" is number of misclassified pixels).}
\label{fig:IllustrateIRW}
\end{figure*}

\begin{figure*}
\centering
\includegraphics[width=6.4in]{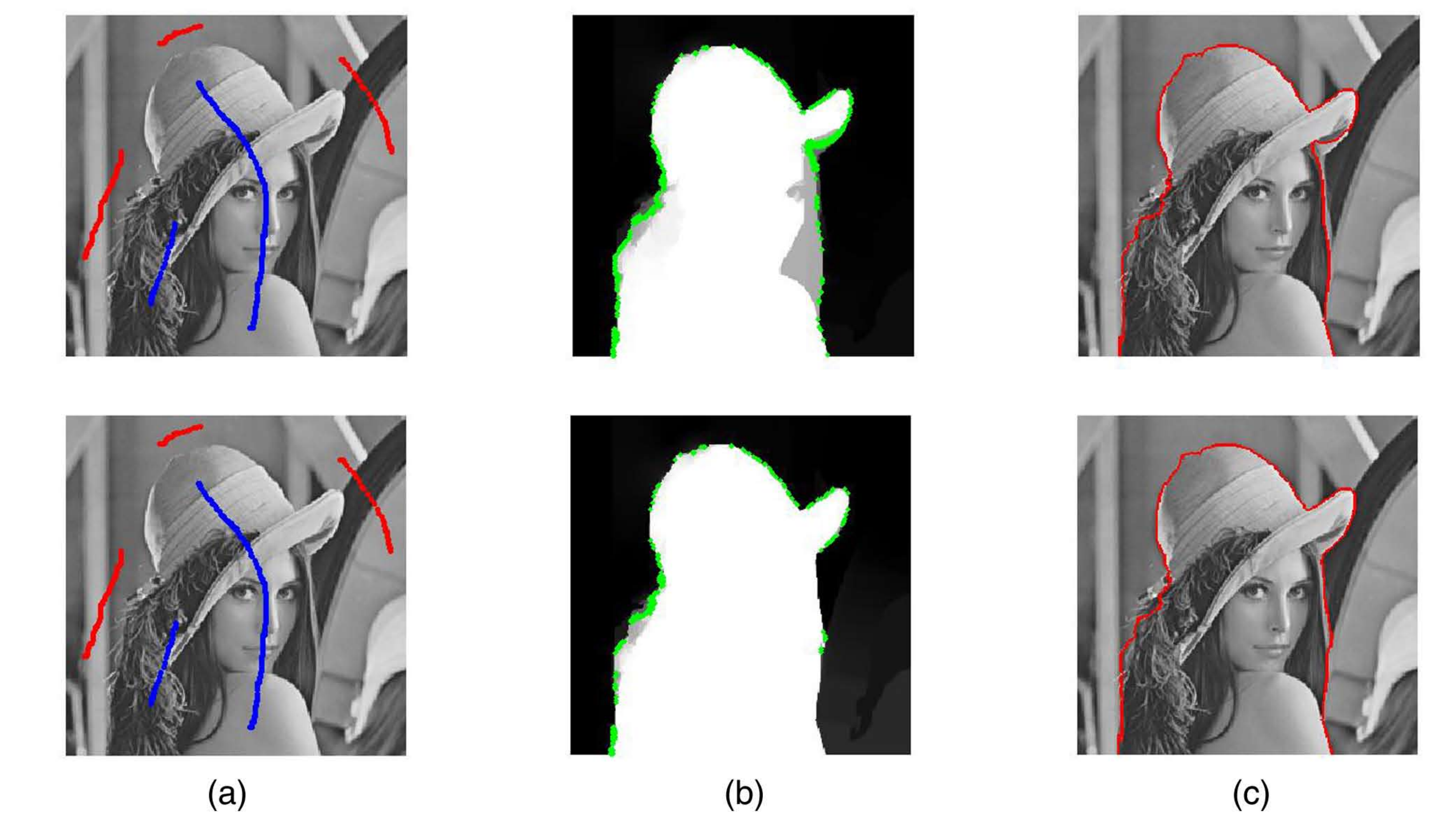}
\caption{Comparison between BRW and random walks algorithm. (a) input images with background and foreground seeds; (b) the probability maps by random walks, and the greed point is boundary pixel whose probability range in $ [ 0.5 - \delta ,0.5 + \delta]$ ($\delta =0.1$); (c) the segmentation results. The top row is of random walks algorithm, and the bottom row is of BRW.}
\label{fig:IllustrateBRW}
\end{figure*}

\begin{figure*}
\centering
\includegraphics[width=6.4in]{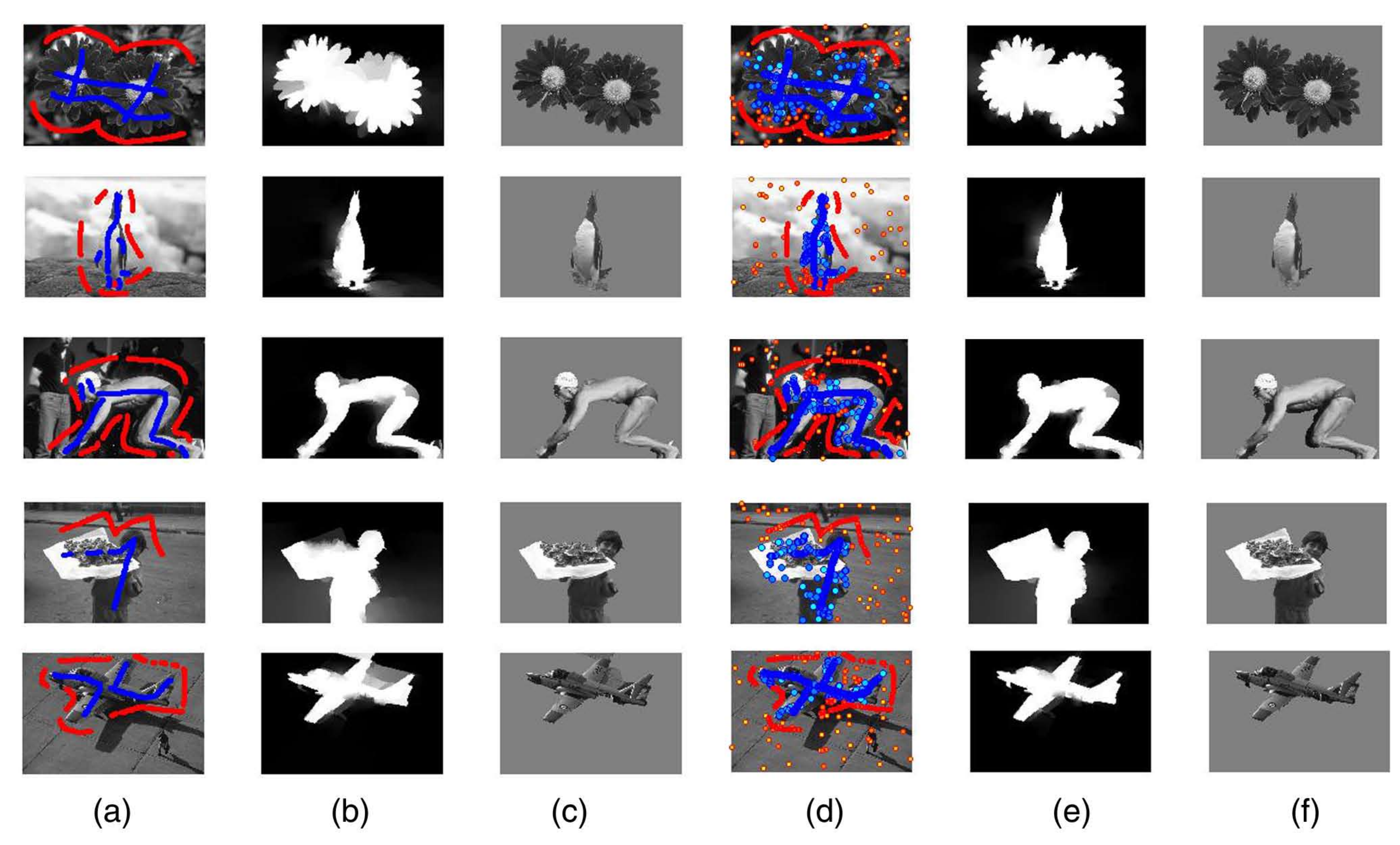}
\caption{Comparison between IBRW and random walks algorithm. From left to right: (a) input images with background and foreground seeds; (b) the probability maps by random walks; (c) the segmentation results by random walks; (d) input images with new background and foreground seeds by IBRW method; (e) the probability maps by IBRW algorithm; (f) segmentation results by IBRW algorithm.}
\label{fig:MoreComparions}
\end{figure*}

We first verify the performance improvement by adding
more background and foreground seeds into random walks
algorithm in each iteration, i.e., the IRW algorithm. The
key merit of IRW is to mitigate the limited training sample
problem by utilizing the previously segmented results. Hence,
it is important to obtain reliable background and foreground
seeds (semi-seeds) based on probability distribution to improve
segmentation results iteratively

Fig. \ref{fig:IllustrateIRW} shows the image segmentation results of IRW algorithm in 3 iterations. The first column of Fig. \ref{fig:IllustrateIRW} is the results of
basic random walks, i.e., the initialization of IRW algorithm.
The next 3 rows are the results of IRW. The IRW automatically
select the semi-seeds by (8). From the second row of Fig. \ref{fig:IllustrateIRW},
i.e., the probability maps of IRW in each iteration, it can see
that the pixels near the left shoulder have higher probability
values after the semi-seeds are involved into the IRW. This
is the positive feedback effect of semi-seeds, i.e., we can find
more semi-seeds from probability maps of lower segmentation
potential, and in return, more semi-seeds lead to probability
distribution of lower segmentation potential. Compared with
the results in Fig. \ref{fig:IllustrateIRW}(a)(d), after adding 70 semi-seeds into
random walk algorithm, the number of misclassified pixels
is reduced from 623 to 584. The boundary pixels whose
probability values close to 0.5 are reduced from 205 to 141.
More details of IRW in segmentation process are shown in the
third and fourth rows of Fig. \ref{fig:IllustrateIRW}. We also provide the zoom-in
segmentation results (left shoulder). From the zoom-in results,
it is clear that the misclassified pixels are iteratively corrected
by adding more semi-seeds. Those experimental results verify
that the introducing of sem-seeds into random walk algorithm
could significantly improve its performance.

\begin{figure}
\centering
\includegraphics[width=3.2in]{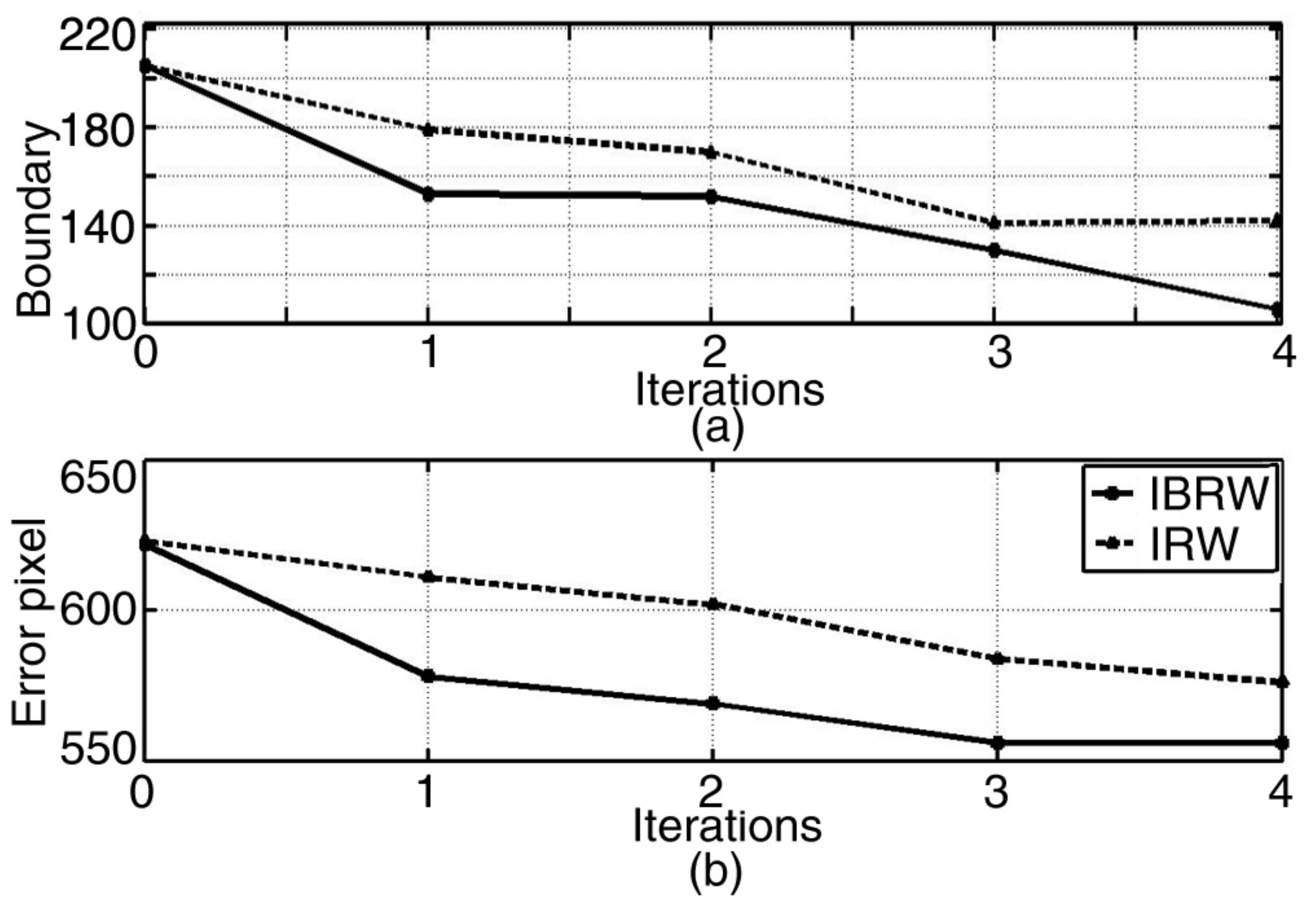}
\caption{Illustration of IBRW v.s. IRW using test image of Fig. \ref{fig:IllustrateIRW}.}
\label{fig:IBRWvsIRW}
\end{figure}

\begin{figure*}
\centering
\includegraphics[width=6.4in]{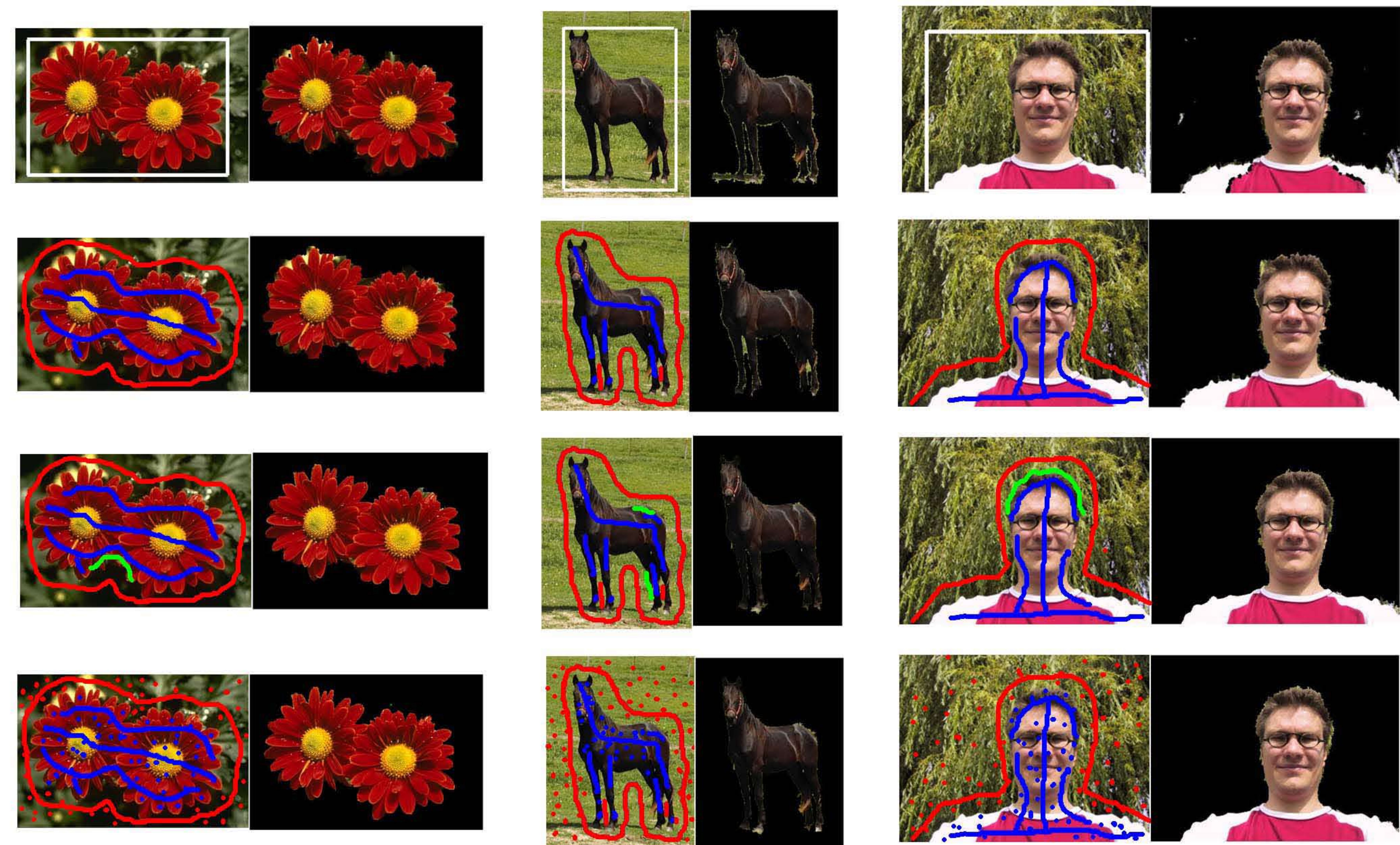}
\caption{Comparing the segmentation results of the GrabCut algorithm, LazySnapping algorithm, constrained random walks algorithm and proposed algorithm. From top to bottom: the results of GrabCut, the results of LazySnapping, the results of constrained random walks, and the results of IBRW. The blue and red line denote foreground and background seeds, especially green line denote soft seeds in constrained random walks.}
\label{fig:SomeAthorAlgorihtms}
\end{figure*}

\subsection{Experimental Study for IBRW Algorithm}


We also carried out some experiments to evaluate the
performance of BRW and IBRW algorithms. In Fig. \ref{fig:IllustrateBRW}, the
BRW is compared with the basic random walk algorithm. To
conduct a fair comparison, same background and foreground
seeds are used to initiate the random walk and BRW algorithm.
From Fig. \ref{fig:IllustrateBRW}(b), it is clear that the number of boundary seeds
is much smaller that that of the random walk algorithm.
The probability map of BRW also has lower segmentation
potential. These experimental results reveal the effectiveness
of the proposed BRW algorithm.

In Fig. \ref{fig:MoreComparions}, more results are presented to compare the performance of IBRW and RW. In the initialization of both algorithms, the same input foreground/background seeds are used
as shown in Fig. \ref{fig:MoreComparions}(a). After several iterations, as shown in Fig. \ref{fig:MoreComparions}(d), the sets of foreground/background seeds for IBRW are
enlarged with adding of semi-seeds which are automatically
selected by IBRW. It can be seen that the segmentation results
of IBRW outperforms the random walks algorithm. Moreover,
from the probability maps of IBRW and RW (Fig. \ref{fig:MoreComparions}(b) and
Fig. \ref{fig:MoreComparions}(e)), it is clear that the segmentation potential has been
efficiently reduced.

In Fig. \ref{fig:IBRWvsIRW}, the performance comparison between IRW and
IBRW is also illustrated in terms of boundary pixels and error
pixels. From the results shown in Fig. \ref{fig:IBRWvsIRW}, it is clear that both
IRW and IBRW have decreased number of boundary and error
pixels in each iteration. Since the IBRW uses the modified
boundary random walk algorithm (10), IBRW has more faster
convergence speed than IRW. From Fig. \ref{fig:IBRWvsIRW}, we can also find
that the IBRW can converge in about 3 iterations generally.

We further compare the performance of IBRW with some
state-of-art algorithms including GrabCut, LazySnapping, constrained random walks on 3 different images. The input of
GrabCut algorithm is a rectangle covering the target object as
shown in the top row of Fig. 8. For the other algorithms, the
same background/foreground strokes are used as inputs. Compared with the other popular image segmentation methods, we can find that proposed IBRW achieves better segmentation
performance.

In the following experiment, 50 images form MSRC
dataset are tested by the proposed IBRW algorithm
and more state-of-art algorithms, like LazySnapping \cite{LazySnapping},
GrabCut \cite{Rother2004grabcut}, SIOX \cite{Friedland2005siox}, Random walks \cite{Grady2006random} and Constrained random walks \cite{Yang2010user}. All of these algorithms use
the exactly same inputs of background/foreground strokes.
Table \ref{tab:EachMethod} shows the error rates of all algorithms. The error
rate is defined as the ratio between number of misclassified
pixels and number of unclassified pixels, where misclassified pixels exclude those from the unclassified region [20].
For the other state-of-art algorithms, we directly quote
the best results from the literature [1], [13]-[16]. It can
be seen that proposed algorithm outperforms the other
state-of-art algorithms and achieves very low error rates.
Meaning that you have different
input for these experiments.

\begin{table}[htbp]
\caption{IBRW compared with other method via MSRC dataset.}
\label{tab:EachMethod}
\centering
\begin{tabular}{|c|c|}
\hline
Method & Error rate\\\hline
LazySnapping \cite{LazySnapping}    & 6.65\%\\\hline
GrabCut \cite{Rother2004grabcut}    & 5.66\%\\\hline
SIOX \cite{Friedland2005siox}       & 9.10\%\\\hline
Random walks \cite{Grady2006random} & 5.40\%\\\hline
Constrained random walks \cite{Yang2010user} & 4.08\%\\\hline
IBRW (proposed method) & 3.85\%\\\hline
\end{tabular}
\end{table}

\subsection{Sensitivity Analysis of Parameters}

\begin{figure}
\centering
\includegraphics[width=3.2in]{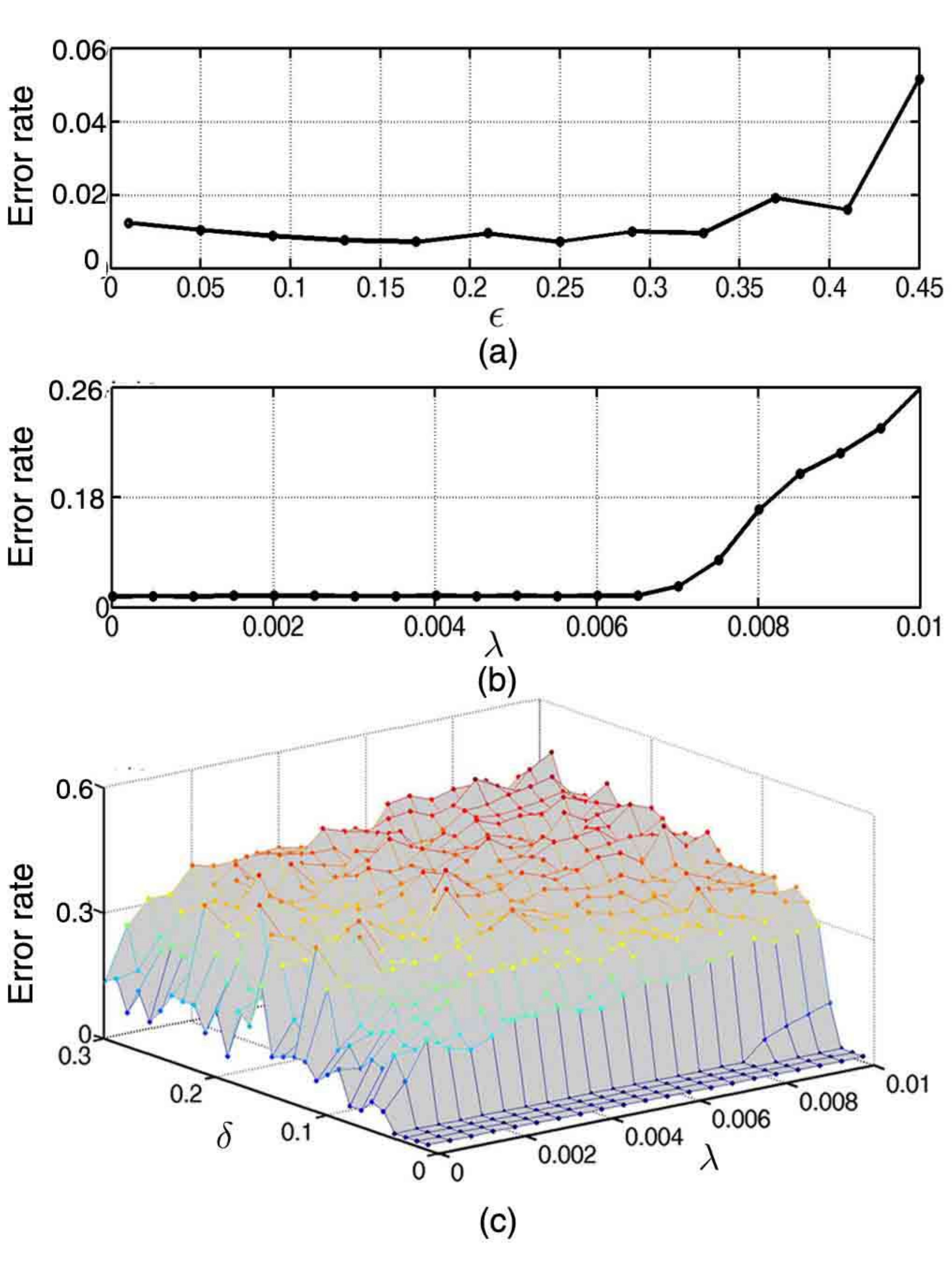}
\caption{Sensitivity of proposed algorithm versus each parameters with Berkeley dataset. a) the threshold $\epsilon$ for selection of new background and foreground seeds; b) the factor $\lambda$ for BRW optimization term; c) combinations of $\lambda$ and $\delta$, especially the parameter $\delta$ decides selection of boundary seeds.}
\label{fig:ParameterSetting}
\end{figure}

There are three parameters used in the proposed algorithm:
the threshold $\epsilon$ for selection of semi-seeds, the parameter $\delta$
for selection of boundary seeds and the trade-off factor $\lambda$
for BRW. It is important to study the sensitivity of proposed
IBRW to these parameters. In this Section, we conduct some
experiments on investigating the sensitivity of the proposed
methods on Berkeley dataset.

The threshold $\epsilon$ plays a critical role in selection of
background/foreground semi-seeds. In the first experiment,
the parameter $\epsilon$ is selected as discrete candidates in range
$[0.01,0.45]$. As shown in Fig. \ref{fig:ParameterSetting}(a), we can find that IRW has
low error rate when the threshold $\epsilon$ in the range of $[0.1, 0.3]$.
However, if the threshold $\epsilon$ is less than 0.4, the error rate is less
sensitive to the parameter. Hence, we suggest to use $\epsilon$ = 0.1
in experiments. If the threshold is too small, the selection of
new seeds will involve in local field. If the threshold
is too big, the selection of new seeds will unreliable.

The trade-off factor $\lambda$ controls the compromise between
the weighted difference of probabilities and the segmentation
potential. We expect that the probabilities of boundary seeds
could be far away 1/2. Hence, it should use larger $\lambda$. However,
to guarantee the problem (10) to be a convex optimization
problem, the factor $\lambda$ is limited in the range $[ 0,\min (w_{ij})]$.
Fig. \ref{fig:ParameterSetting} (b) shows that the IBRW algorithm performance is
deteriorate when the factor $\lambda$ is out of range $[ 0,\min (w_{ij})]$.

Lastly, we evaluate the sensitivity of IBRW versus different
combinations of $\lambda$ and $\delta$. Generally, large value of $\delta$ is
encouraged to emphasize the selection of boundary seeds.
The IBRW algorithm achieves bad performance for large
δ values, since many non-boundary pixels are regarded as
boundary seeds. As shown in Fig. \ref{fig:ParameterSetting} (c), there are many
parameter combination near the optimal performance on
area of small λ and δ. We fixed $(\lambda,\delta)$ as (0.005, 0.1) in
our experiments.

\section{Conclusion}
The interactive image segmentation algorithm under limited
user input play an important role in industrial application.
It can avoid the expensive and tedious user input. To let
machine more intelligent understand the intention of limited
user input, the pixels of self-learning must present on the
image segmentation. In this study, the most important work
is to reveal positive feedback system on image segmentation.
It provides a way to extend the basic random walks algorithm.
We use the iterative boundary random walks algorithms, which
it is combined iterative random walks with boundary random
walks approach, to reduce segmentation potential. Experiment
results on image segmentation shows that proposed algorithms
can obtain more efficient input. And higher segmentation
performance can be obtained by applying the IBRW algorithm.
The proposed methods can also be applied to many other
pattern recognitions field, e.g. limited training sample. Future
work could focus on the selection of semi-seeds, instead of
the threshold method in IBRW.

\ifCLASSOPTIONcaptionsoff
  \newpage
\fi

\end{document}